\documentclass[letterpaper]{article} 
\usepackage{aaai23}  
\usepackage{times}  
\usepackage{helvet}  
\usepackage{courier}  
\usepackage[hyphens]{url}  
\usepackage{graphicx} 
\usepackage{natbib}  
\usepackage{caption} 
\usepackage{algorithm}
\usepackage{algorithmic}
\usepackage{enumerate} 
\usepackage{multirow}

\newcommand{\eat}[1]{}
\usepackage{color}
\usepackage{url}
\usepackage{amsmath}
\usepackage{amsthm}
\usepackage{amssymb}
\usepackage{bm}
\usepackage{newfloat}
\usepackage{listings}
\DeclareCaptionStyle{ruled}{labelfont=normalfont,labelsep=colon,strut=off} 
\floatstyle{ruled}
\newfloat{listing}{tb}{lst}{}
\floatname{listing}{Listing}
\nocopyright

\title{RepParser: End-to-End Multiple Human Parsing with Representative Parts}
\author{
Xiaojia Chen,
Xuanhan Wang,
Lianli Gao,
Jingkuan Song
}
\affiliations{
{\rm Center for Future Media,\\
	University of Electronic Science and Technology of China}	
}

\usepackage{bibentry}

\begin{document}

\maketitle

\begin{abstract}
Existing methods of multiple human parsing usually adopt a two-stage strategy (typically top-down and bottom-up), which suffers from either strong dependence on prior detection or highly computational redundancy during post-grouping. 
In this work, we present an end-to-end multiple human parsing framework using representative parts, termed \textbf{\textit{RepParser}}. 
Different from mainstream methods, RepParser solves the multiple human parsing in a new single-stage manner without resorting to person detection or post-grouping.
To this end, RepParser decouples the parsing pipeline into \textbf{\textit{instance-aware kernel generation}} and \textbf{\textit{part-aware human parsing}}, which are responsible for instance separation and instance-specific part segmentation, respectively. 
In particular, we empower the parsing pipeline by \textbf{\textit{representative parts}}, since they are characterized by instance-aware keypoints and can be utilized to dynamically parse each person instance. 
Specifically, representative parts are obtained by jointly localizing centers of instances and estimating keypoints of body part regions. 
After that, we dynamically predict instance-aware convolution kernels through representative parts, thus encoding person-part context into each kernel responsible for casting an image feature as an instance-specific representation.
Furthermore, a multi-branch structure is adopted to divide each instance-specific representation into several part-aware representations for separate part segmentation.
In this way, RepParser accordingly focuses on person instances with the guidance of representative parts and directly outputs parsing results for each person instance, thus eliminating the requirement of the prior detection or post-grouping.
Extensive experiments on two challenging benchmarks demonstrate that our proposed RepParser is a simple yet effective framework and achieves very competitive performance. We show that it significantly outperforms most two-stage methods and variants of single-stage instance recognition methods. 
\end{abstract}

\begin{figure}[h!]

\begin{center}
	\includegraphics[width=0.95\linewidth,height=1.1\linewidth]{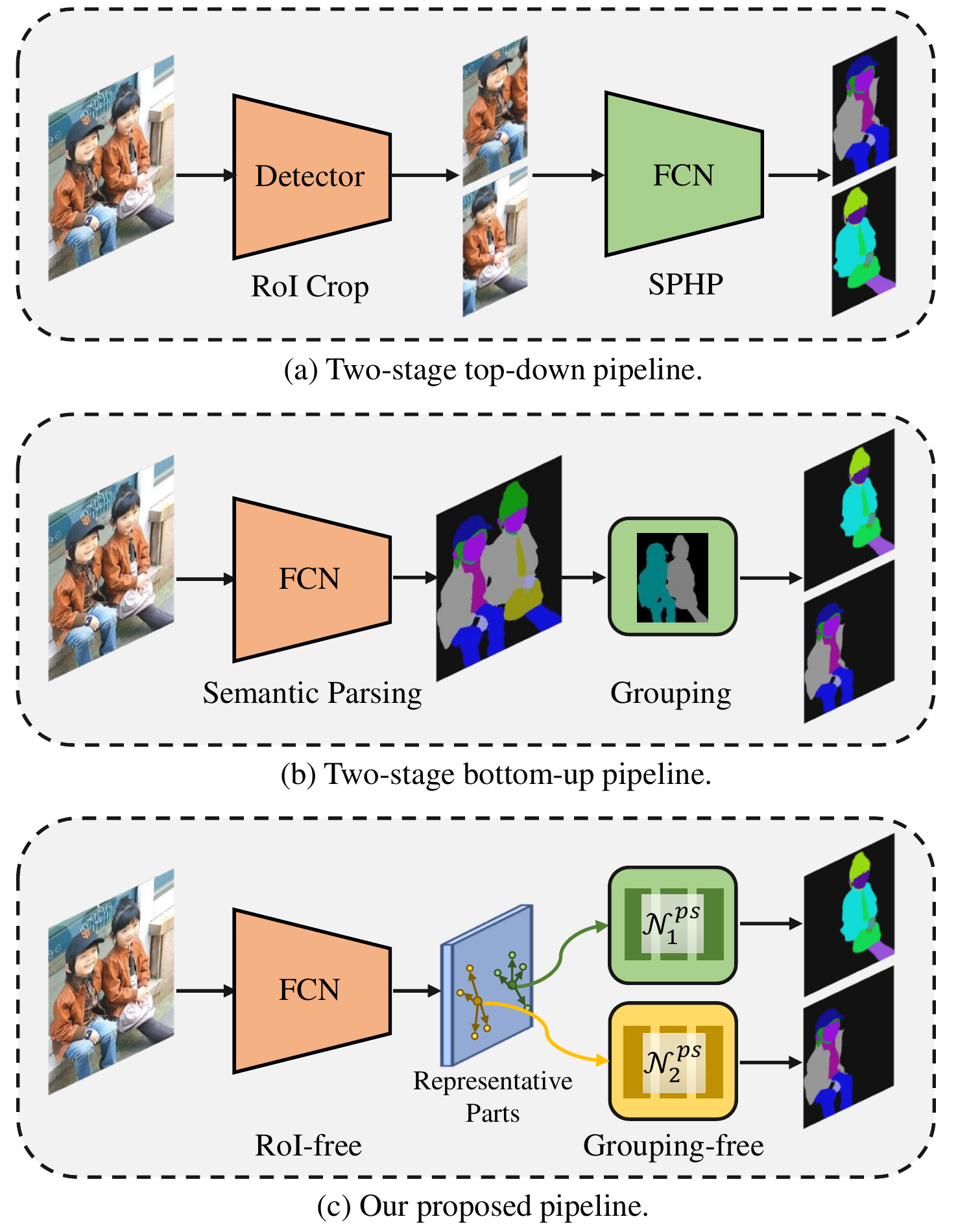}
\end{center}
\caption{Different multiple human parsing pipelines: SPHP denotes single-person human parsing. $N_{i}^{ps}$ indicates the parsing network for $i$-th person. }
\label{fig:motivation}

\end{figure}
\section{Introduction} 

Multiple human parsing (MHP) aims to segment body parts for each person in an image, which is a fundamental yet challenging task in the human-centric intelligence system. Compared to many dense predictions tasks, such as object detection~\cite{obj_det:faster} and instance segmentation \cite{ins_seg:mask_rcnn,inst_condinst,panoptic_FCN,yu2022soit}, it is the arbitrary number of fine-grained body parts that have made MHP much more challenging. 
In particular, the success of MHP models depends on two key aspects: 1) whether the model can make a correct instance separation, and 2) whether this model can decide what semantic behind each image pixel.
Inspired by the success of the human-centric recognition, such as pose estimation \cite{pose:stip,WangKTN++,pose_SunXLWang2019,WangAMA++}, existing methods of multiple human parsing adopt a two-stage strategy, which consists of top-down and bottom-up pipelines. In particular, the top-down pipeline (Fig.~\ref{fig:motivation}.(a)) starts with person detection responsible for the first aspect, then an RoI operation is adopted to crop the person from the feature maps or the original image. After that, single-person human parsing is performed for addressing the second aspect. Instead, the bottom-up pipeline (Fig.~\ref{fig:motivation}.(b)) firstly segments instance-agnostic body parts responsible for the second aspect, then groups them into instance-aware results for addressing the first aspect. Despite the great progress, previous state-of-the-art methods for multiple human parsing still encounter several challenges, as analyzed below:

The strong coupling of the second stage with the first stage in the two-stage framework significantly hampers high-quality multiple human parsing. Specifically, top-down methods \cite{parsing:semtree,densepose:parsingrcnn,ins_seg:mask_rcnn,yang2020eccv} are highly dependent on person detection results while bottom-up methods rely on instance-agnostic part segmentation results. Since person bounding boxes are rectangular, they may contain irrelevant contents such as body parts belonging to other persons. In this way, the performance of human parsing will drop significantly if the person detection performance decreases dramatically. In terms of bottom-up pipeline, existing methods \cite{seg_part:cihp,Parsing:Graphonomy,parsing:mhpwild,seg_part:mhp} predict redundant instance-agnostic body parts. In this way, some body parts may be removed during grouping post-processing due to their low-quality confidential scores. Besides, the processing of assembling body parts (e.g., Hungary algorithm) is often heuristic, making these methods complicated yet inefficient. Overall, the bottleneck of the two-stage frameworks lies in the first stage, as the performance of a model in the first stage decides the upper bound of the entire algorithm. 


The above challenges motivate us to rethink two problems: 1) how to design a single-stage pipeline for multiple human parsing, and 2) how to equip this pipeline with the ability to establish a direct mapping from an image to various instance-specific body parts. 
To handle the above two problems, we present an end-to-end multiple human parsing framework using representative parts, termed \textbf{\textit{RepParser}}. As illustrated in Fig.~\ref{fig:motivation}.(c), the proposed RepParser is designed in an end-to-end manner without resorting to person detection or post-grouping. 
To this end, RepParser decouples the parsing pipeline into \textbf{\textit{instance-aware kernel generation}} and \textbf{\textit{part-aware human parsing}}, which are responsible for instance separation and instance-specific part segmentation, respectively.
The core idea is that we empower the parsing pipeline with representative parts, since they are characterized by instance-aware keypoints and can be utilized to dynamically parse each person instance. 
Specifically, representative parts are obtained by jointly localizing centers of instances and estimating keypoints of body part regions. 
After that, we dynamically predict instance-aware convolution kernels through representative parts, thus encoding person-part context into each kernel responsible for casting an image feature as an instance-specific representation.
Furthermore, a multi-branch structure is adopted to divide each instance-specific representation into several part-aware representations for separate part segmentation.
In this way, RepParser accordingly focuses on person instances with the guidance of representative parts and directly outputs parsing results for each person instance, eliminating the need for person detection or body part grouping. In summary, our work has the following contributions:
\begin{enumerate}[(1)]
\item We propose a novel multiple human parsing pipeline termed \textit{RepParser}, which eliminates the dependence of prior person detection and avoids heuristic post-grouping operations.   
\item The RepParser is designed in a flexible fashion, as it dynamically encodes person-part contexts into corresponding convolution kernels. To our knowledge, this is the first single-stage method for multiple human parsing and it can inspire related research on fine-grained recognition.
\item  Extensive experiments conducted on two challenging benchmarks demonstrate the effectiveness and generalizability of the proposed method. Moreover, it significantly outperforms most two-stage methods and variants of single-stage instance recognition methods.
\end{enumerate}
\begin{figure*}
\centering
\includegraphics[width=0.9\linewidth,height=0.45\linewidth]{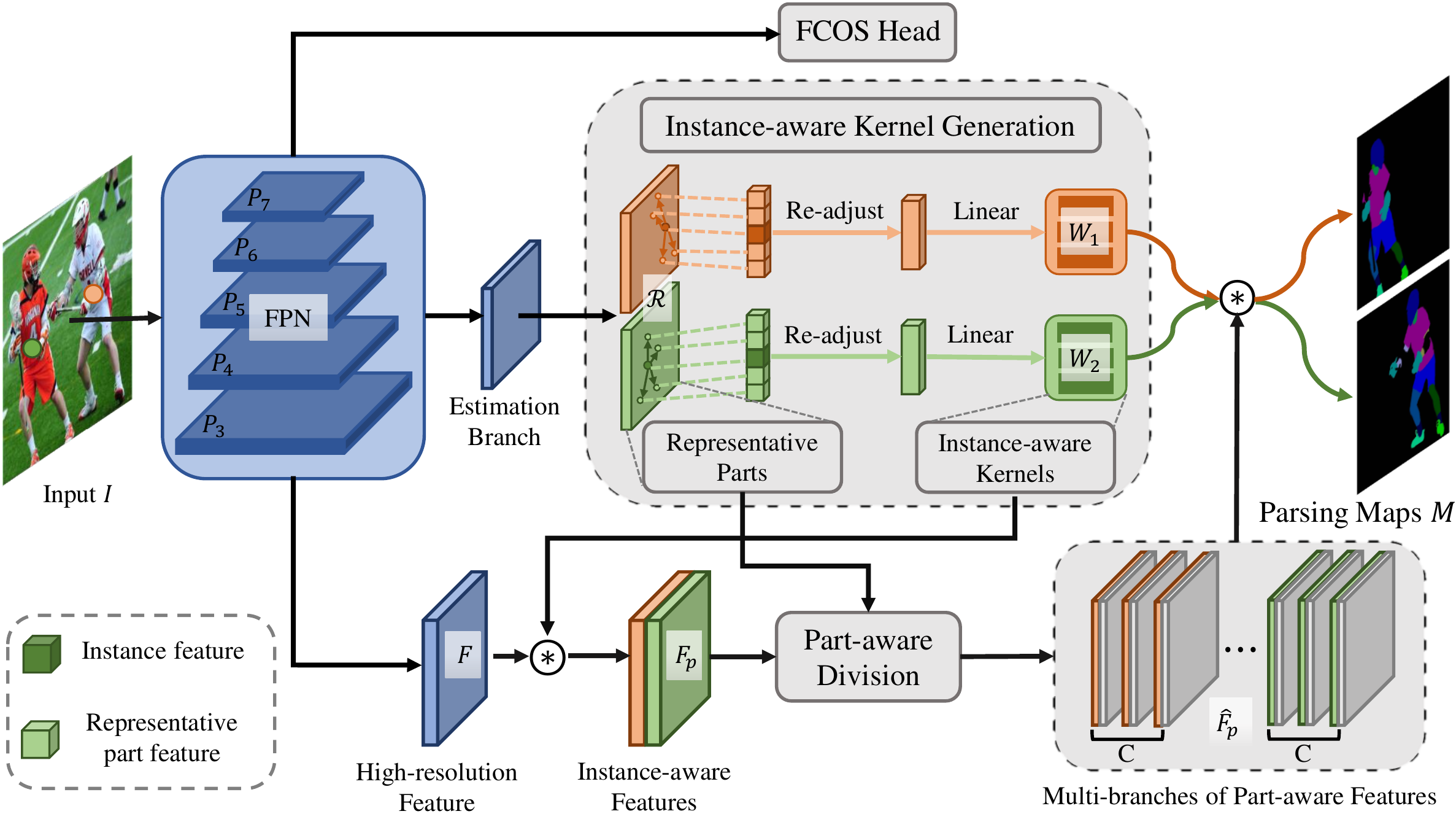}
\caption{The overall architecture of RepParser without resorting to prior detection or post-grouping, where it decouples the multiple human parsing pipeline into instance-aware kernel generation and part-aware human parsing. In particular, it firstly estimates several representative parts, which are dynamically responsible for instance-aware feature generation. Then, a multi-branch structure is adopted to divide instance-aware features into part-aware features for separate part segmentation.}
\label{fig:network}

\end{figure*}

\section{Related Work}
\subsection{Multiple Human Parsing}
To date, methods of multiple human parsing are based on the two-stage pipeline. Most of them can be divided into two categories: 1) bottom-up paradigm and 2) top-down paradigm. As mentioned above, the bottom-up methods \cite{seg_part:cihp,Parsing:Graphonomy,parsing:mhpwild,seg_part:mhp} regard multiple human parsing as a segment-then-grouping pipeline. The series of the bottom-up methods usually generate redundant human parsing results, leading to high computational costs during post-processing. 
Compared with bottom-up series, top-down approaches \cite{parsing:semtree,densepose:parsingrcnn,ins_seg:mask_rcnn,yang2020eccv,he2021progressive,Liu_Zhang_Yang_Yuen_2021} focus on the single-person human parsing problem as they employ person detector to solve the issue of person separation. Furthermore, recent works have developed two versions of top-down framework: the unified top-down model \cite{densepose:parsingrcnn,ins_seg:mask_rcnn,yang2020eccv,parsing:unified} and the separated top-down model \cite{parsing:semtree,parsing:mce2p,parsing:braidnet}. The difference between the two versions is whether unifying the person detector into a single-person parsing model. For example, Mask-RCNN~\cite{ins_seg:mask_rcnn} can be regarded as the first unified top-down approach, where it adopts Faster R-CNN \cite{obj_det:faster} to predict a bounding box for each person and extracts region-of-person from detector's features for performing instance-specific part segmentation. Following this idea, Parsing R-CNN \cite{densepose:parsingrcnn} and RP R-CNN \cite{yang2020eccv} devote to solving the problem of the single-person human parsing and propose new variants of Mask R-CNN via contextual modeling or part re-scoring. 

Different from two-stage methods, our work is devoted to designing a novel single-stage pipeline and focuses on instance-aware body part segmentation with representative parts.

\subsection{Single-stage Instance-level Recognition}
Traditional solutions try to build an instance-specific model for instance-level recognition. For example, Tian \textit{et al.} \cite{inst_condinst} adopt conditional convolutions for one-stage instance segmentation, where each convolution kernel is dynamically generated from a center point of person instance. This design improves the instance segmentation performance while maintaining high efficiency. Moreover, Li \textit{et al.} \cite{panoptic_FCN} propose a location-aware kernel generation for panoptic scene understanding.
Mao~\textit{et al.} \cite{Mao2021pose} propose to dynamically generate a keypoint-aware estimator for multi-person pose estimation.
Although these different approaches vary in tasks, they all share a characteristic: they focus on the instance-specific convolution kernel generation.
However, each convolution kernel generated through existing methods encodes sparse content of an instance (i.e., object center only). Therefore, the generated kernels severely ignore the person-part context which is essential for accurate human parsing, thus leading to suboptimal results as demonstrated in the experimental results.  

As a supplement to them, we extend instance-specific modeling to solve the multiple human parsing. 
Instead of directly deriving from single-stage frameworks that are used in other instance recognition tasks, we propose to parse multiple human instances through representative parts, and encode person-part context into each instance-aware convolution kernels as well as part representation. As a result, the proposed method significantly outperforms variants of single-stage methods applied in other instance recognition tasks.

\section{Methodology}
The pipeline of our RepParser is presented in Fig.~\ref{fig:network}. Given an input image $\mathit{I}$, the goal of multiple human parsing is to localize $N$ person instances and segment $C$ body parts for each localized person. In particular, it needs to address two issues: 1) how to distinguish each person instance from other instances or background; and 2) how to perform instance-aware parsing without extra operations (\textit{i.e.}, RoI cropping or part grouping). To address these, we propose to parse multiple persons using representative parts. Specifically, RepParser firstly utilizes a backbone network (\textit{e.g.}, ResNet) to obtain an image-level feature $F$ with a size of $H\times W\times D$, where $D$ indicates the number of channels and $\{H, W\}$ denotes the spatial size. Next, a detection branch, which is an FCOS~\cite{tian2019fcos} head with an object center estimator and a location regressor, is adopted to localize person instances. With the location of person centers, an instance-aware kernel generation branch is used to estimate the representative parts of each person and accordingly generate convolution kernels for each instance. With instance-aware kernels and representative parts, a part-aware parsing module, which is a multi-branch structure, is utilized to generate part-aware features for accurate human parsing. In the following, we describe the details of RepParser.

\subsection{Representative Parts}
As discussed before, bounding boxes and object centers are often used to represent person instances in two-stage methods and single-stage methods. Due to the rectangular shape, a bounding box has a rough global context of a person but cannot account for the semantically important local areas. Instead, the object center only accounts for small local areas, thus ignoring the interrelation among body parts and the global context of an instance. 
To overcome these limitations, our core idea is that each person instance in an image is represented by representative parts. It is expected that the representative parts can encode the characteristics of each person instance and only focus on the pixels of corresponding body parts.
Motivated by this, we propose to dynamically construct representative parts of an instance through keypoints of body parts, as they can reflect the global context of a person (\textit{e.g.}, posture or shape) and semantically salient part areas.
Formally, let $\mathcal{R} = \{(x_k, y_k)\}_{k=1}^C$ denotes representative parts of a person, where $(x_k, y_k)$ is the keypoint of the $k$-th part  (\textit{e.g.}, face, left-arm, right-arm, and so on), $C$ is the number of part categories (\textit{e.g.}, $C$=20 for CIHP dataset). Thus, we parse person instances conditioning on their representative parts, as they not only present characteristics of pose and shape but also reflect person-part relations.

To construct representations of representative parts, we need to localize them by object centers. 
As shown in Fig.~\ref{fig:network}, we firstly adopt a feature pyramid network~\cite{lin2017feature} to produce multi-scale feature maps from levels 3 to 7. Following FCOS~\cite{tian2019fcos}, we treat each location on the feature maps as a potential instance. Thus, for each location $(x_h, y_h)$ on the feature maps, we estimate the confidential score being a person center and the offsets to representative parts. Based on this, representative parts $\mathcal{R}$ is calculated by Eq.~\ref{equ.partloc}.
\begin{equation}
\mathcal{R} = \{(x_h + \Delta x_k, y_h + \Delta y_k)\}_{k=1}^C,
\label{equ.partloc}
\end{equation}
where $\{(\Delta x_k, \Delta y_k)\}_{k=1}^C$ are the normalized offsets from the center of a person instance to the center of the representative parts. After that, we construct an initial representation of representative parts by sampling pixel points from the image-level feature $F$. Formally, we denote $f_h$ as the feature of sampled instance point and $\{f_h^k\}_{k=1}^C$ as the features of representative parts. Next, we employ these sampled representative parts for instance-aware kernel generation and part-aware human parsing.

\subsection{Instance-aware Kernel Generation}
To obtain high-quality instance representations for accurately human parsing, it is expected that the instance-aware convolution kernels are dynamically generated by relying on the characteristics of instances. To achieve this, we propose to generate instance-aware kernels by representative parts, since they encode potential contexts about person-part relations. 
Instead of directly applying initial representative parts to predict instance-aware kernels, we first re-adjust the representation of the representative parts according to the person-part relations, aiming to dynamically encode the person-part context into the corresponding kernel. Specifically, the re-adjusted representative parts are obtained through Eq.~\ref{equ.context_gate}:
\begin{equation}
\begin{array}{lll}
\alpha & = \{\sigma(W_a[f_h \oplus f_h^k])\}_{k=1}^C, &   \\
f_p & = W_m[(\alpha_{1}\cdot f_h^1) \oplus...\oplus (\alpha_{C}\cdot f_h^C)], &   \\
\end{array}
\label{equ.context_gate} 
\end{equation}
where $W_a$ and $W_m$ are learnable parameters that are respectively responsible for relation estimation and feature re-adjustment. $\sigma(\cdot)$ is the standard sigmoid function and $\oplus$ means a concatenation operation. The $\alpha$ is the estimated relation matrix, where each element in $\alpha$ denotes a part being relevant to a person with a confidential score. The $f_p$ denotes the representation of representative parts, which is dynamically generated via person-part interaction. Note that the estimated relational score $\alpha$ is continually updated by directly supervised training so that it becomes more accurate from time to time. 

Given re-adjusted representative parts $f_p$ and instance representation $f_h$, we generate two types of convolution kernels through Eq.~\ref{equ.inst_kernel}. 
\begin{equation}
\begin{array}{lll}
W_f &  = V_1(f_h \oplus f_p), &   \\
W_o &  = V_2(f_h \oplus f_p), &   \\
\end{array}
\label{equ.inst_kernel}
\end{equation}
where $V_1$ and $V_2$ are linear matrices for kernel generation. $W_f$ is used to project an image feature to an instance-aware feature without resolution reduction, while the $W_o$ is responsible for predicting part masks from many part-aware features. Notably, the generated kernels are very compact, as they are designed in a $1\times1$ convolution layer with less channels (e.g., 32 for $W_f$ and C for $W_o$). We project the image-level feature $F$ to the instance-aware feature for human parsing by Eq.~\ref{equ.inst_conv}.
\begin{equation}
\begin{array}{lll}
F_p &  = W_f \circledast F, &   \\
\end{array}
\label{equ.inst_conv}
\end{equation}
where $\circledast$ is the convolution operation. Compared with top-down methods, such as Parsing-RCNN consisting of eight $3\times3$ convolution layers with 256 channels for instance feature extraction, the generated kernels are much more lightweight.

\subsection{Part-aware Human Parsing}
After handling the issue of person separation by representative parts, we would like to predict an accurate mask for each part. Instead of predicting masks from the instance-aware feature, we first build multiple branches for separate parsing. In each branch, we construct a part-aware representation that can be used to only fire on the pixels of the corresponding part. Specifically, we first divide the instance-aware feature $F_p$ into $C$ groups, each of which is responsible for part-specific segmentation. For each group, we construct a geometry map that records relative distances from all pixels to the corresponding representative part, suggesting the salient area of the corresponding part. Formally, we denote all geometry maps as $F_s$ and compute the part-aware representation by Eq.~\ref{equ.inst_part}.
\begin{equation}
\begin{array}{lll}
\hat{F_p} &  = \{W_p^k[F_p^k \oplus F_s^k]\}_{k=1}^{C}, &   \\
\end{array}
\label{equ.inst_part}
\end{equation}
where $W_p$ is the learnable transformation matrix. Next, we utilized dynamically generated kernel $W_o$ to predict mask for each part, which is formalized by Eq.~\ref{equ.output}.
\begin{equation}
\begin{array}{lll}
M &  = \{\phi(W_o^k \circledast \hat{F_p^k})\}_{k=1}^{C}, &   \\
\end{array}
\label{equ.output}
\end{equation}
where $M$ is the predicted parsing maps and $\phi(\cdot)$ is the standard softmax function. Although it is possible to predict masks by instance-aware feature, we empirically find that using part-aware representation performs better.

\begin{table*}[t]
\centering
\renewcommand\arraystretch{1.1}
\resizebox{0.99\linewidth}{!}{
	\begin{tabular}{|c|cccc|cccc|c}
		\hline
		\multicolumn{1}{c|}{Method}       & \multicolumn{1}{c|}{Backbone}   & \multicolumn{1}{c|}{Epoch} &  \multicolumn{1}{c}{RoI-free} & \multicolumn{1}{c|}{Grouping-free}  & \multicolumn{1}{c|}{mIoU} & \multicolumn{1}{c|}{${\rm{AP}}^p_{vol}$} & \multicolumn{1}{c|}{${\rm{AP}}_{50}^p$} & ${\rm{PCP}}_{50}$ & Time (ms)\\ \hline

		\multicolumn{10}{c}{Bottom-up}\\ \hline
		\multicolumn{1}{c|}{PGN~\cite{seg_part:cihp}}           & \multicolumn{1}{c|}{-} & \multicolumn{1}{c|}{-}   & \checkmark  &               & \textbf{25.3} & 35.5 & 17.6      & 26.9  &  497   \\ 
		
		\multicolumn{1}{c|}{MH-Parser~\cite{parsing:mhpwild}}        & \multicolumn{1}{c|}{ResNet-101} & \multicolumn{1}{c|}{-}  & \checkmark  &             & - & 36.0    & 17.9         & 26.9      & 1486    \\ 
		
		\multicolumn{1}{c|}{NAN~\cite{seg_part:mhp}}        & \multicolumn{1}{c|}{-} & \multicolumn{1}{c|}{80}   & \checkmark  &              & - & \textbf{41.7}    & \textbf{25.1}  & \textbf{32.2} & 1037  \\ \hline
		
		\multicolumn{10}{c}{Top-down} \\ \hline 
		
		\multicolumn{1}{c|}{Mask RCNN~\cite{ins_seg:mask_rcnn}}     & \multicolumn{1}{c|}{ResNet-50}  & \multicolumn{1}{c|}{-}  &  & \checkmark                  & - & 33.9 & 14.9      & 25.1 & 243 ($\ddagger$)  \\
		
		\multicolumn{1}{c|}{Parsing RCNN~\cite{densepose:parsingrcnn}}  & \multicolumn{1}{c|}{ResNet-50}  & \multicolumn{1}{c|}{25}  &  & \checkmark                 & 34.0 & 36.7 & 19.9      & 32.4   & 270 ($\intercal$)    \\ 
		
		\multicolumn{1}{c|}{Parsing RCNN~\cite{densepose:parsingrcnn}}  & \multicolumn{1}{c|}{ResNet-50}  & \multicolumn{1}{c|}{75}  &  & \checkmark                 & 36.1 & 40.5 & 27.4      & 38.3   & 270   \\ 
		
		\multicolumn{1}{c|}{SemaTree~\cite{parsing:semtree}}       & \multicolumn{1}{c|}{ResNet-101} & \multicolumn{1}{c|}{200}  &  & \checkmark                & - & 42.5 & 34.4      & 43.5    & 3234      \\ 
		
		\multicolumn{1}{c|}{M-CE2P~\cite{parsing:mce2p}}       & \multicolumn{1}{c|}{ResNet-101} & \multicolumn{1}{c|}{150} &  & \checkmark                 & \textbf{41.1} & 42.7 & 34.5      & \textbf{43.8}     & 1107     \\ 
		
		\multicolumn{1}{c|}{RP-RCNN~\cite{yang2020eccv}}  & \multicolumn{1}{c|}{ResNet-50}  & \multicolumn{1}{c|}{75}  &  & \checkmark                & 37.3 & \textbf{45.2} & \textbf{40.5}      & 39.2    & 394 ($\diamond$)   \\ \hline
		
		\multicolumn{10}{c}{Single-stage}\\ \hline
		\multicolumn{1}{c|}{DETR$^{\ast}$~\cite{obj_det:detr}} & \multicolumn{1}{c|}{ResNet-50}  & \multicolumn{1}{c|}{25}   & \checkmark & \checkmark                & 30.5 & 33.7 & 12.1      & 25.1     & 218 ($\ddagger$) (\textcolor{red}{$\downarrow 10\%$})  \\ 
		\multicolumn{1}{c|}{Deformable DETR$^{\ast}$~\cite{obj_det:deformdetr}} & \multicolumn{1}{c|}{ResNet-50}  & \multicolumn{1}{c|}{25}   & \checkmark & \checkmark                & \textbf{33.4} & 34.8 & 14.2     & 29.4   &  241 ($\intercal$) (\textcolor{red}{$\downarrow 11\%$})   \\ 
		\multicolumn{1}{c|}{condInst$^{\ast}$~\cite{inst_condinst}} & \multicolumn{1}{c|}{ResNet-50}  & \multicolumn{1}{c|}{25}   & \checkmark & \checkmark                & 26.2 & \textbf{36.5} & \textbf{18.7}      & \textbf{30.1}    & 164 ($\intercal$) (\textcolor{red}{$\downarrow 39\%$})   \\ \hline
		
		\multicolumn{1}{c|}{RepParser (Ours)} & \multicolumn{1}{c|}{ResNet-50}  & \multicolumn{1}{c|}{25}   & \checkmark & \checkmark                & 35.9 & 39.4 & 25.5      & 36.8   & 193 ($\intercal$) (\textcolor{red}{$\downarrow 29\%$})    \\ 
		
		\multicolumn{1}{c|}{RepParser (Ours)} & \multicolumn{1}{c|}{ResNet-50}  & \multicolumn{1}{c|}{75}   & \checkmark & \checkmark                & 38.3 & 42.3 & 33.7      & 43.4   & 193 ($\diamond$)  (\textcolor{red}{$\downarrow 51\%$})    \\  
		
		\multicolumn{1}{c|}{RepParser (Ours)} & \multicolumn{1}{c|}{ResNet-101}   & \multicolumn{1}{c|}{75}   & \checkmark & \checkmark                & 39.7 & 43.0 & 35.6      & \textbf{45.2}  & 208 ($\diamond$) (\textcolor{red}{$\downarrow 47\%$})   \\ 
		
		\multicolumn{1}{c|}{RepParser (Ours)} & \multicolumn{1}{c|}{Swin-S}     & \multicolumn{1}{c|}{75}   & \checkmark & \checkmark               & \textbf{41.1} & \textbf{45.6} & \textbf{42.4}      & \textbf{55.0}  & 220 ($\diamond$)  (\textcolor{red}{$\downarrow 44\%$})    \\ 
		
		\hline
	\end{tabular}
}
\caption{Comparison with state-of-the-art methods on MHP-v2 validation set. The symbol ``$\ast$'' means that model is a re-implemented version. In addition to time costs, the relative proportion of time costs reduction caused by single-stage models is also reported. A single-stage model and a two-stage model are marked with the same symbol, as they achieve comparable parsing performance but with different time costs. The RepParser with ResNet-50 backbone achieves competitive results to the best competitor RP-RCNN~\cite{yang2020eccv}, with much fewer time costs.}
\label{table:mhp}

\end{table*}

\begin{table*}[t]
\centering
\renewcommand\arraystretch{1.1}
\resizebox{0.99\linewidth}{!}{
	\begin{tabular}{c|cccc|cccc|c}
		\hline
		\multicolumn{1}{c|}{Method}       & \multicolumn{1}{c|}{Backbone}   & \multicolumn{1}{c|}{Epoch}
		& \multicolumn{1}{c}{RoI-free} & \multicolumn{1}{c|}{Grouping-free}
		& \multicolumn{1}{c|}{mIoU} & \multicolumn{1}{c|}{${\rm{AP}}^p_{vol}$} & \multicolumn{1}{c|}{${\rm{AP}}_{50}^p$} & ${\rm{PCP}}_{50}$ & Time (ms)\\ \hline
		
		\multicolumn{10}{c}{Bottom-up} \\ \hline
		
		\multicolumn{1}{c|}{PGN~\cite{seg_part:cihp}}           & \multicolumn{1}{c|}{ResNet-101} & \multicolumn{1}{c|}{80}    
		& \checkmark  &               & 55.8 & 39.0 & 34.0      & 61.0  & 497     \\ 
		
		\multicolumn{1}{c|}{Graphonomy~\cite{Parsing:Graphonomy}}        & \multicolumn{1}{c|}{Xception} & \multicolumn{1}{c|}{100}    
		& \checkmark &   & \textbf{58.6} & -    & -         & -    & -      \\ \hline
		
		\multicolumn{10}{c}{Top-down} \\ \hline 
		
		\multicolumn{1}{c|}{Mask RCNN~\cite{ins_seg:mask_rcnn}}     & \multicolumn{1}{c|}{ResNet-50}  & \multicolumn{1}{c|}{25}     &
		&     \checkmark
		& 47.7 & 45.2 & 42.0      & 44.0    &  243 ($\ddagger$) \\ 
		
		\multicolumn{1}{c|}{Mask RCNN~\cite{ins_seg:mask_rcnn}}     & \multicolumn{1}{c|}{ResNet-50}  & \multicolumn{1}{c|}{75}     &
		&     \checkmark
		& 51.1 & 47.4 & 49.4      & 49.5   &  243   \\ 
		
		\multicolumn{1}{c|}{Parsing RCNN~\cite{densepose:parsingrcnn}}  & \multicolumn{1}{c|}{ResNet-50}  & \multicolumn{1}{c|}{25}     &
		&     \checkmark
		& 52.8 & 51.2 & 57.2      & 55.4   &  270 ($\intercal$)  \\ 
		\multicolumn{1}{c|}{Parsing RCNN~\cite{densepose:parsingrcnn}}  & \multicolumn{1}{c|}{ResNet-50}  & \multicolumn{1}{c|}{75}     &
		&     \checkmark
		& 56.3 & 53.9 & 63.7      & 60.1   &   270 \\ 
		\multicolumn{1}{c|}{Unified~\cite{parsing:unified}}       & \multicolumn{1}{c|}{ResNet-101} & \multicolumn{1}{c|}{37}     &
		&     \checkmark
		& 55.2 & 48.0 & 51.0      & -  &  -\\ 
		
		\multicolumn{1}{c|}{M-CE2P~\cite{parsing:mce2p}}       & \multicolumn{1}{c|}{ResNet-101} & \multicolumn{1}{c|}{200}     &
		&     \checkmark
		& 59.5 & - & -      & -    &   1107   \\ 
		
		\multicolumn{1}{c|}{BraidNet~\cite{parsing:braidnet}}       & \multicolumn{1}{c|}{ResNet-101} & \multicolumn{1}{c|}{150}     &
		&     \checkmark
		& 60.6 & - & -     & -    &   -   \\
		
		\multicolumn{1}{c|}{SemaTree~\cite{parsing:semtree}}       & \multicolumn{1}{c|}{ResNet-101} & \multicolumn{1}{c|}{200}     &
		&    \checkmark
		& \textbf{60.9} & - & -     & -    &  3234    \\ 
		\multicolumn{1}{c|}{RP-RCNN~\cite{yang2020eccv}}  & \multicolumn{1}{c|}{ResNet-50}  & \multicolumn{1}{c|}{75}     &
		&  \checkmark  
		& 58.2 & \textbf{58.3} & \textbf{71.6}      & \textbf{62.2}   &  394 ($\diamond$)  \\ \hline
		
		\multicolumn{10}{c}{Single-stage}\\ \hline 
		\multicolumn{1}{c|}{DETR$^{\ast}$~\cite{obj_det:detr}} & \multicolumn{1}{c|}{ResNet-50}  & \multicolumn{1}{c|}{25}   & \checkmark & \checkmark                & 48.3 & 43.8 & 39.3      & 44.2    & 218 ($\ddagger$) (\textcolor{red}{$\downarrow 10\%$})   \\ 
		\multicolumn{1}{c|}{Deformable DETR$^{\ast}$~\cite{obj_det:deformdetr}} & \multicolumn{1}{c|}{ResNet-50}  & \multicolumn{1}{c|}{25}   & \checkmark & \checkmark                & 46.4 & 44.0 & 38.5      & 44.0     & 241 ($\intercal$) (\textcolor{red}{$\downarrow 11\%$})  \\ 
		\multicolumn{1}{c|}{condInst$^{\ast}$~\cite{inst_condinst}} & \multicolumn{1}{c|}{ResNet-50}  & \multicolumn{1}{c|}{25}   & \checkmark & \checkmark                & \textbf{49.7} & \textbf{47.1} & \textbf{46.9}      & \textbf{48.1}    &  164 ($\intercal$) (\textcolor{red}{$\downarrow 39\%$})  \\ \hline
		\multicolumn{1}{c|}{RepParser (Ours)} & \multicolumn{1}{c|}{ResNet-50}  & \multicolumn{1}{c|}{25}     
		&  \checkmark  & \checkmark
		& 52.9 & 51.9 & 57.5      & 55.7  &  193 ($\intercal$) (\textcolor{red}{$\downarrow 29\%$})    \\ 
		
		\multicolumn{1}{c|}{RepParser (Ours)} & \multicolumn{1}{c|}{ResNet-50}   & \multicolumn{1}{c|}{75}     
		&  \checkmark  & \checkmark
		& 56.3 & 53.1 & 61.5      & 59.3     & 193 ($\diamond$)  (\textcolor{red}{$\downarrow 51\%$})  \\ 
		
		\multicolumn{1}{c|}{RepParser (Ours)} & \multicolumn{1}{c|}{ResNet-101}   & \multicolumn{1}{c|}{75}   & \checkmark & \checkmark                & 57.9 & 54.4 & 64.9     & 61.5  & 208 ($\diamond$) (\textcolor{red}{$\downarrow 47\%$})   \\ 
		
		\multicolumn{1}{c|}{RepParser (Ours)} & \multicolumn{1}{c|}{Swin-S}     & \multicolumn{1}{c|}{75}     
		&  \checkmark  & \checkmark
		& \textbf{61.7} & \textbf{57.2} & \textbf{70.4}     & \textbf{65.8}     & 220 ($\diamond$)  (\textcolor{red}{$\downarrow 44\%$})  \\
		
		\hline 
	\end{tabular}
}
\caption{Comparison with state-of-the-art methods on CIHP validation set. The symbol ``$\ast$'' means that model is a re-implemented version.}
\label{table:cihp}

\end{table*}

\section{Experiments}
\subsection{Experimental Setup}
\subsubsection{Datasets:}
Our experiments are conducted on two challenging multiple human parsing datasets:  MHP-v2~\cite{seg_part:mhp} and CIHP~\cite{seg_part:cihp}. MHP-v2 is a commonly used dataset for instance-level human parsing. It is split into 15k/5k/5k images for train/val/test. Each image contains an average of three people with 58 body part categories.  In addition, the CIHP dataset is currently the largest multiple human parsing dataset, which covers 19 part categories and involves many crowded scenes. It is split into 28k/5k/5k images for train/val/test.
\subsubsection{Metrics:}
For evaluation, we use many standard metrics to measure the performance of all parsing models, including the \textbf{A}verage \textbf{P}recision based on \textbf{p}art (${\rm{AP}}^p$) and \textbf{P}ercentage of \textbf{C}orrectly parsed semantic \textbf{P}arts ($\rm{PCP}$). We report ${\rm{AP}}^p_{vol}$ and ${\rm{AP}}^p_{50}$. ${\rm{AP}}^p_{vol}$ is the average of ${\rm{AP}}^p$ at different IoU thresholds ranging from 0.1 to 0.9. In particular, ${\rm{AP}}^p_{50}$ means that ${\rm{AP}}^p$ is calculated at an IoU threshold of 0.5. In terms of instance-agnostic parsing, we report \textbf{m}ean \textbf{I}ntersection-\textbf{O}ver-\textbf{U}nion (mIOU) for model evaluation.
\subsubsection{Implementation details:}
Our RepParser is implemented based on MMDetection~\cite{chen2019mmdetection} on eight NVIDIA Tesla V100 GPUs. Following FCOS~\cite{tian2019fcos}, FPN~\cite{lin2017feature} is used as the feature extraction network. The weights of all backbones are pre-trained on ImageNet, while the remaining weights are randomly initialized. We train models using SGD/AdamW for convolution-/transformer-based models, respectively. A mini-batch size of 16 is used. Other details are identical to FCOS~\cite{tian2019fcos}.

\subsection{Main Results}

In this section, we compare proposed RepParser with state-of-the-art multiple human parsing methods and report evaluation results summarized from two datasets. In addition to existing two-stage methods that are either based on top-down paradigm or bottom-up paradigm, we also compare RepParser with other representative single-stage methods that are applied for other instance recognition tasks, including DETR~\cite{obj_det:detr}, Deformable DETR~\cite{obj_det:deformdetr} and condInst~\cite{inst_condinst}. Moreover, we re-implement these single-stage methods and train them under the same settings for a fair comparison, since these methods are not evaluated for multiple human parsing in original papers. In addition to standard metrics, we also measure inference time per image for each method on the same hardware if possible. Furthermore, we also report the relative proportion of time costs reduction for investigation of the efficiency of each single-stage model.

\begin{figure}[t]
\centering
\includegraphics[width=0.9\linewidth,height=0.5\linewidth]{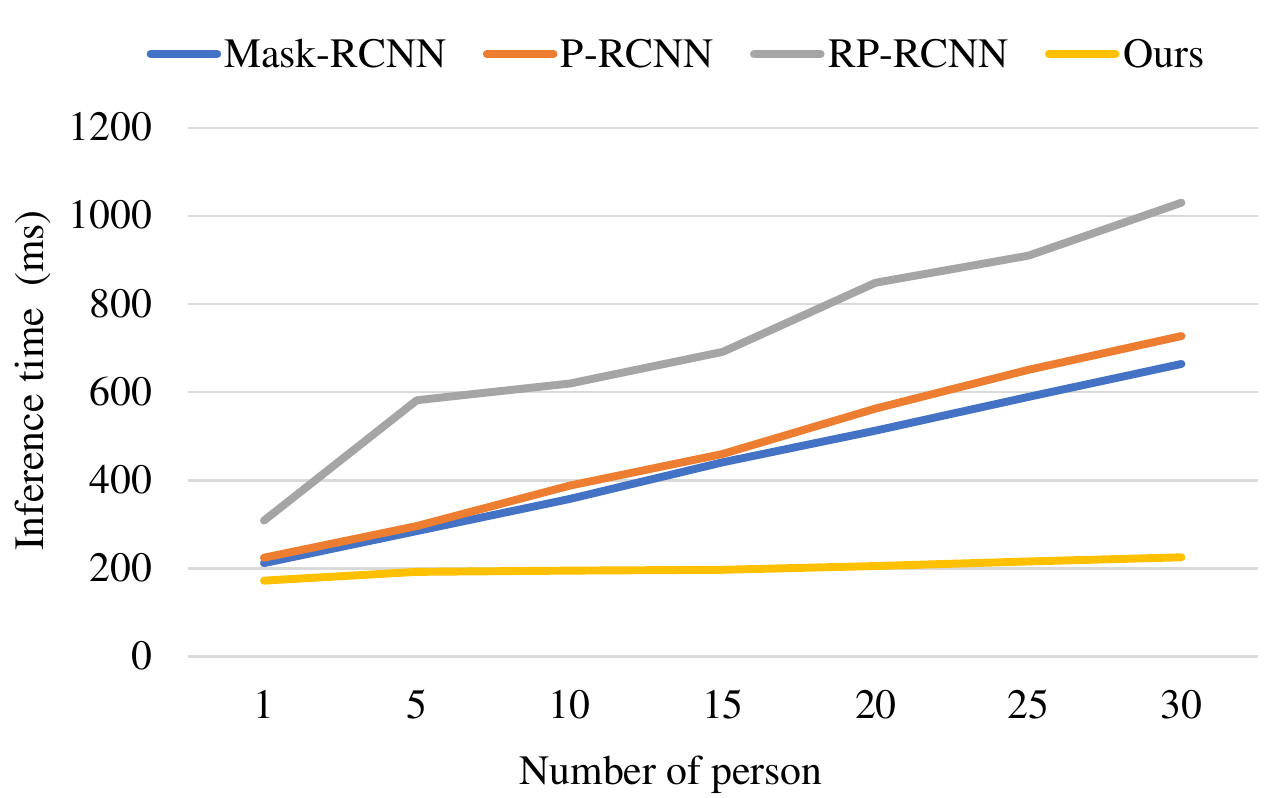}
\caption{The inference time w.r.t the number of the persons in an image.}
\label{fig:compare}

\end{figure}
\begin{figure}[t]
\centering
\includegraphics[width=0.9\linewidth]{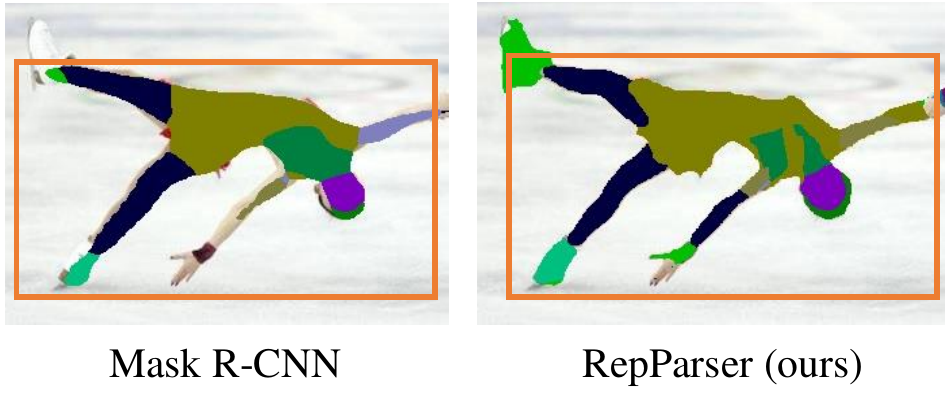}
\caption{Comparison with RoI-based method: The Mask-RCNN cannot handle a part outside a box but RepParser can break this limitation.}
\label{fig:visual}

\end{figure}

\begin{figure*}[t]
\centering
\includegraphics[width=0.8\linewidth,height=0.4\linewidth]{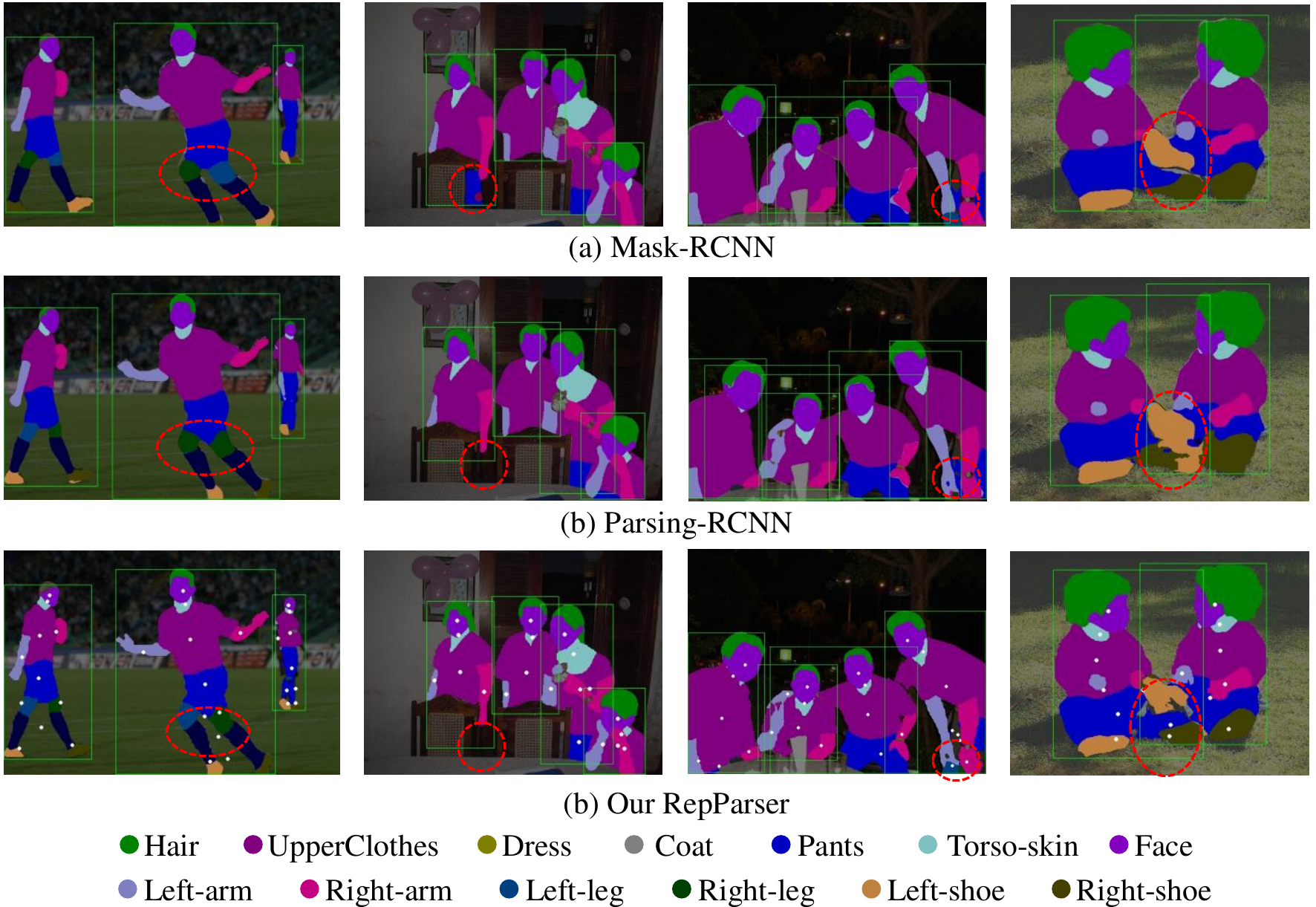}
\caption{Qualitative comparison. From top to bottom: the parsing results obtained from Mask-RCNN \cite{ins_seg:mask_rcnn}, PR-RCNN \cite{yang2020eccv} and our RepParser. The red circles spot the difference between the two models. The white dots are estimated representative parts by RepParser.}
\label{fig:result}

\end{figure*}
\noindent\textbf{MHP-v2:}
As shown in Table~\ref{table:mhp}, we evaluate RepParser on MHP-v2 validation set and compare it with state-of-the-art multiple human parsing methods. 
From the results, we find that RepParser achieves very competitive parsing results, which are comparable to or higher than that of state-of-the-art methods. 
Compared with the previous bottom-up methods, RepParser with ResNet-50 backbone has better performances (38.3\% mIoU vs 25.3\% mIoU, 42.3\% ${\rm{AP}}^p_{vol}$ vs 41.7\% ${\rm{AP}}^p_{vol}$) with much fewer time costs (193ms vs 1037ms).

As for the comparison with top-down methods, RepParser achieves competitive parsing performance with fewer time costs. For example, RepParser with ResNet-50 significantly outperforms the Mask-RCNN under the same settings, e.g., 35.9\% mIoU vs 33.4\% mIoU, 39.4\% ${\rm{AP}}^p_{vol}$ vs 33.9\% ${\rm{AP}}^p_{vol}$ and 36.8\% ${\rm{PCP}}_{50}$ vs 26.8\% ${\rm{PCP}}_{50}$. On two stronger methods using M-CE2P and RP-RCNN, RepParser shows comparable parsing results under the same setting, but has much lower time costs (\textit{e.g.}, 208ms vs 1107ms, 193ms vs 394ms). 
It is worth noting that many non-unified top-down methods such as SemaTree~\cite{parsing:semtree} adopt an isolated object detector to detect persons and then crop the RoIs on the original images, thus leading to high inference costs (\textit{i.e.}, from the input image to the parsing results).
As illustrated in Fig.~\ref{fig:compare}, the inference time of top-down methods, such as Mask RCNN, Parsing RCNN, and PR-RCNN, dramatically increases as the number of persons linearly increases. 
However, the RepParser keeps almost constant inference time, since it eliminates the prior detection and each generated kernel is very compact (\textit{i.e.}, $32\times32\times1\times1$). 
This suggests that the RepParser could be applied to many complex real-world scenes, such as crowded scene, while keeping stable yet high efficiency.
Moreover, another major merit of the RepParser is that it does not rely on bounding boxes. As illustrated in Fig.~\ref{fig:visual}(left), top-down methods only perform human parsing inside the predicted bounding boxes. As a result, some body parts cannot be parsed if the detector yield inaccurate bounding boxes. However, RepParser can well handle the body parts even outside the box (see Fig.~\ref{fig:visual}(right)).

In terms of the comparison with single-stage parsing methods, the RepParser achieves higher parsing performance than that of other single-stage methods, while maintaining competitive efficiency.
This indicates that directly deriving single-stage methods from other instance tasks leads to suboptimal results since they severely ignore the instance-part contexts that are essential for accurate multiple human parsing.

\noindent\textbf{CIHP:}
Similar to experiments conducted on MHP-v2 dataset, we compare RepParser with state-of-the-art methods on CIHP validation set. Corresponding results are listed in Table~\ref{table:cihp}. 
In line with findings from Tab.~\ref{table:mhp}, RepParser also achieves competitive performance on CIHP validation set. For example, RepParser with ResNet-50 backbone has a better performance than that of bottom-up models and significantly outperforms other single-stage models under the same setting. 
Moreover, it also performs comparable to the best top-down competitor RP-RCNN and outperforms other top-down competitors by a margin, but requires less computational costs. Some qualitative results are shown in Fig.~\ref{fig:result}, which clearly demonstrates the effectiveness of our proposed method.

\begin{table}[t]
\centering
\renewcommand\arraystretch{1.1}
\resizebox{0.95\linewidth}{!}{
	\begin{tabular}{cccccc}
		\hline
		baseline &  KG    & PF & mIoU  & ${\rm{AP}}^p_{vol}$  & $\rm{PCP}_{50}$ \\ \hline
		\checkmark	&            &            & 49.7  & 47.1                      & 48.1    \\ 
		\checkmark	& \checkmark &            & 52.5  & 50.7                      & 54.0    \\ 
		\checkmark	& \checkmark & \checkmark & 52.9  & 51.9                      & 55.7  \\ \hline
\end{tabular}}
\caption{Ablation study on representative parts. KG means kernel generation using representative parts. PF means the part-aware feature generation using representative parts.}
\label{table:relation}

\end{table}

\subsection{Ablation Experiments} 
\begin{table}[t]
\centering
\renewcommand\arraystretch{1.1}
\resizebox{0.8\linewidth}{!}{
	\begin{tabular}{ccccc}
		\hline
		width & mIoU  & ${\rm{AP}}^p_{vol}$ & ${\rm{AP}}^p_{50}$ & $\rm{PCP}_{50}$ \\ \hline
		8        & 52.5 & 51.3                  & 56.4   & 54.4    \\
		16       &  53.0 &        51.6          &  57.3  & 55.4    \\
		32       & 52.9 & \textbf{51.9}                  & \textbf{57.5}   & \textbf{55.7}    \\
		64       & \textbf{53.4} & 51.5                  & 57.1   & 55.3    \\ \hline
		\hline
		depth   & mIoU  & ${\rm{AP}}^p_{vol}$   & ${\rm{AP}}^p_{50}$ & $\rm{PCP}_{50}$ \\ \hline
		2        & 52.9 & \textbf{51.9}                  & \textbf{57.5}   & \textbf{55.7}    \\
		3        & \textbf{53.1} & 51.7                  & 57.1   & 55.4    \\
		4        & 52.7 & 51.3                  & 56.4   & 54.9    \\ \hline
\end{tabular}}
\caption{Investigating the effect of kernel scale.}
\label{table:channel}

\end{table}
\noindent\textbf{The effect of representative parts.}
As discussed before, the representative parts separately contribute to instance-aware kernel generation and part-aware feature generation. Thus, we choose the condInst \cite{inst_condinst} with ResNet-50 backbone as the baseline and gradually incorporate representative parts into the pipeline. The experimental results are summarized in Table \ref{table:relation}. From the results, we have the following observations: First, compared with baseline, applying representative parts to predict convolution kernels leads to a significant improvement, achieving 2.8\% of mIoU, 3.6\% of ${\rm{AP}}^p_{vol}$ and 5.9\% of $\rm{PCP}_{50}$ higher than that of baseline. This indicates that encoding person-part context into convolution kernel is particularly important for instance-aware feature generation. Second, constructing part-aware representation via representative parts brings a stable improvement, e.g., improving the score of ${\rm{AP}}^p_{vol}$ from 50.7\% to 51.9\%. This suggests that focusing on salient areas derived from representative parts is particularly beneficial for accurate human parsing. 

\noindent\textbf{The effect of the kernel scale.}
In this section, we investigate the effect of the kernel scale. Here, we consider two factors: the width of each generated convolution kernel and the depth of generated convolution kernels. Our baseline consists of two 1x1 convolutions with 32 channels and performs convolution on the 1/8 down-sampling ratio feature maps. We conduct experiments by adjusting the number of channels or varying the number of convolution layers. As reported in Table \ref{table:channel}, the performance improves as the width increases, but it seems to be saturated when the width is set as 32. However, increasing depth has a negligible effect on parsing performance. Thus, one can conclude that simply enlarging model capacity has reached the performance bottleneck.

\noindent\textbf{Qualitative results.}
As shown in Fig.~\ref{fig:result}, RepParser can produce good parsing results which are comparable to those of two-stage methods. Furthermore, two-stage methods fail to handle identical parts appearing in an intersection of two bounding boxes (see column 4). In contrast, this has a minor effect on RepParser, as it does not rely on bounding boxes. On the other hand, estimated representative parts tend to be located on semantic parts of persons, thus benefiting the instance-aware human parsing. For more details, we refer the reader to supplementary materials.
\section{Conclusion}
In this paper, we propose a new single-stage multiple human parsing method termed RepParser, aiming at breaking the limitation of the two-stage pipeline. To achieve this goal, we utilize representative parts to generate instance-aware kernels as well as part-aware representations, thus facilitating instance-aware human parsing. Extensive experiments on two benchmarks prove the effectiveness of our method.

\bibliography{aaai23}
\bibliographystyle{aaai23}
\clearpage
\section{Appendix A: Implementation Details.}
In this section, we provide more details about implementations, including the effect of the feature resolution and the details of training schedule.

\subsection{The effect of the feature resolution}
Many works \cite{pose_SunXLWang2019,WangAMA++} have demonstrated that higher resolution representation brings better performance for dense prediction tasks. Inspired by this, we investigate which level of the image feature is beneficial for human parsing. Hence, we separately apply generated instance-aware kernels on three different feature maps, which separately are 1/16, 1/8, and 1/4 smaller than the size of the image. Table \ref{table:ratio} indicates that the performance drops dramatically if the resolution of the input feature map is downsampled to 1/16 of the input image. We conjecture the possible reason behind this is that the human parsing task requires pixel-level understanding and high-resolution feature maps preserve more visual contents. However, larger resolution leads to a high computation burden. Besides, generating parsing results from the image feature at the 1/4 scale of the image size brings minor gains, when compared it with the counterpart at 1/8 scale of the image size. Thus, we choose 1/8 as the default setting for a better trade-off between accuracy and speed.

\begin{table}[h]
\centering
\renewcommand\arraystretch{1.3}
\resizebox{0.9\linewidth}{!}{
	\begin{tabular}{cccccc}
		\hline
		Ratio & mIoU  & ${\rm{AP}}^p_{vol}$ & ${\rm{AP}}^p_{50}$ & $\rm{PCP}_{50}$ \\ \hline
		1/16  & 51.9 & 49.2                  & 51.8   & 52.4    \\ 
		1/8   & 52.9 & \textbf{51.9}                  & \textbf{57.5}   & \textbf{55.7}\\
		1/4   & \textbf{53.5} & 51.6                  & 57.3   & 55.7    \\ \hline
\end{tabular}}
\caption{Ablation study on CIHP val with different resolutions of the input feature maps. 'Ratio' denotes the down-sampling ratio of the input feature maps}
\label{table:ratio}

\end{table}

\subsection{Details of training schedule}
In general, a good initialization of models leads to better performance. Thus we explore the impact of initialization methods on multiple human parsing. As shown in Tab~\ref{table:init}, human parsing methods pre-trained on the COCO keypoint dataset can improve ${\rm{AP}}_{vol}^p$ by 1\%. It indicates that a good initialization will lead to better parsing results. 

\begin{table}[h]
\centering
\renewcommand\arraystretch{1.3}
\resizebox{1.0\linewidth}{!}{
	\begin{tabular}{cccccc}
		\hline
		Initialization & mIoU  & ${\rm{AP}}^p_{vol}$ & ${\rm{AP}}^p_{50}$ & $\rm{PCP}_{50}$ \\ \hline
		ImageNet & 52.9 & 51.9 & 57.5 & 55.7    \\ 
		COCO     & 54.1 & 52.8 & 59.8 & 57.4    \\ \hline
\end{tabular}}
\caption{Ablation study on CIHP val. Investigating the way of initialization.}
\label{table:init}
\vspace{-1em}
\end{table}

\section{Appendix B: More Qualitative Results}
In this section, we provide additional qualitative results of our RepParser on CIHP \textit{val} set, including estimated representative parts and failure cases.
\begin{figure*}[t]
\centering
\includegraphics[width=0.95\linewidth]{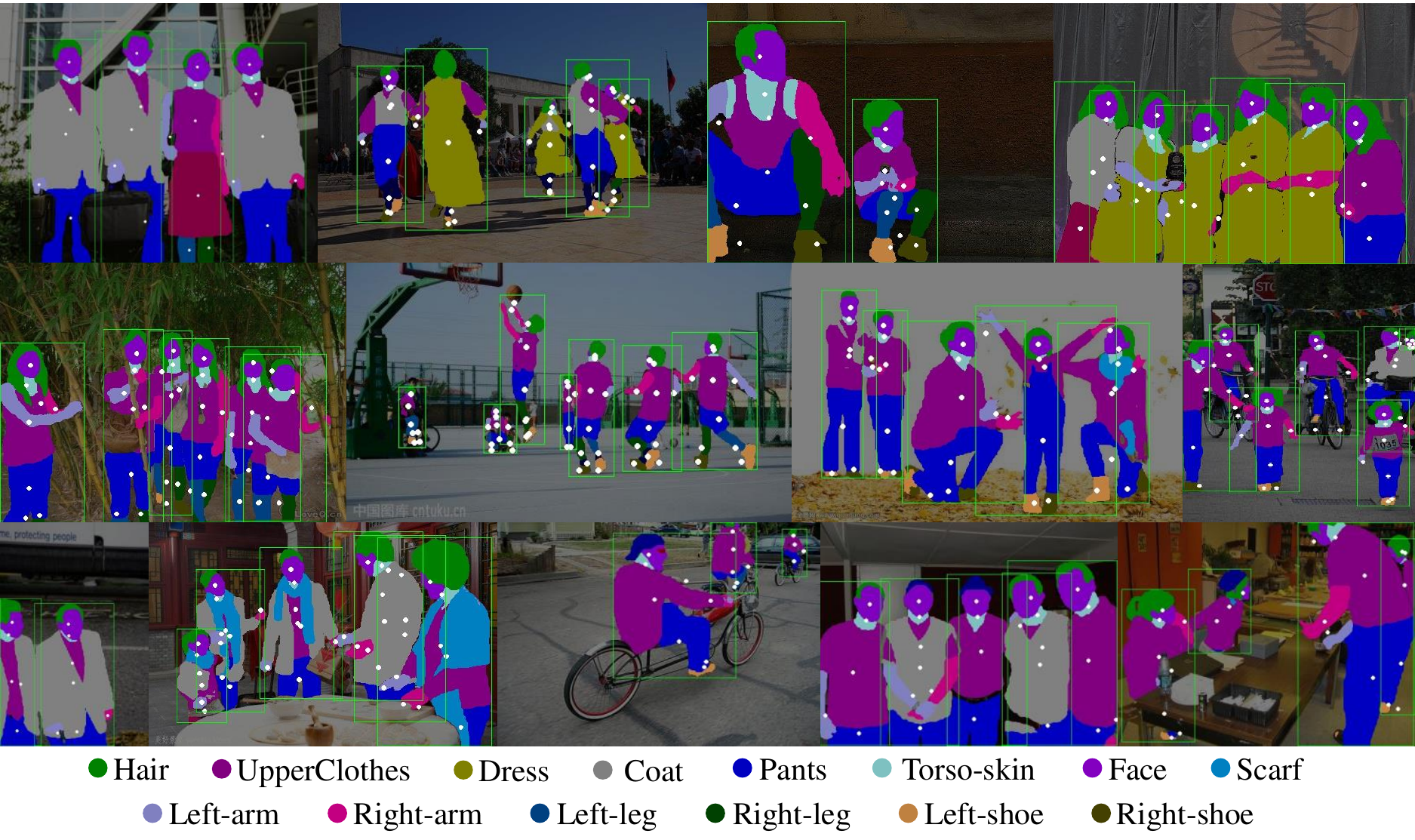}
\caption{More qualitative results of the proposed RepParser on CIHP \textit{val}. PerParser can well handle many challenging scenes with occlusions, scale variations, \textit{etc}. The white dots are estimated representative parts by RepParser. Zoom in for a better view.}
\label{fig:quality_result}
\end{figure*}

\begin{figure*}[t!]
\centering
\includegraphics[width=0.95\linewidth]{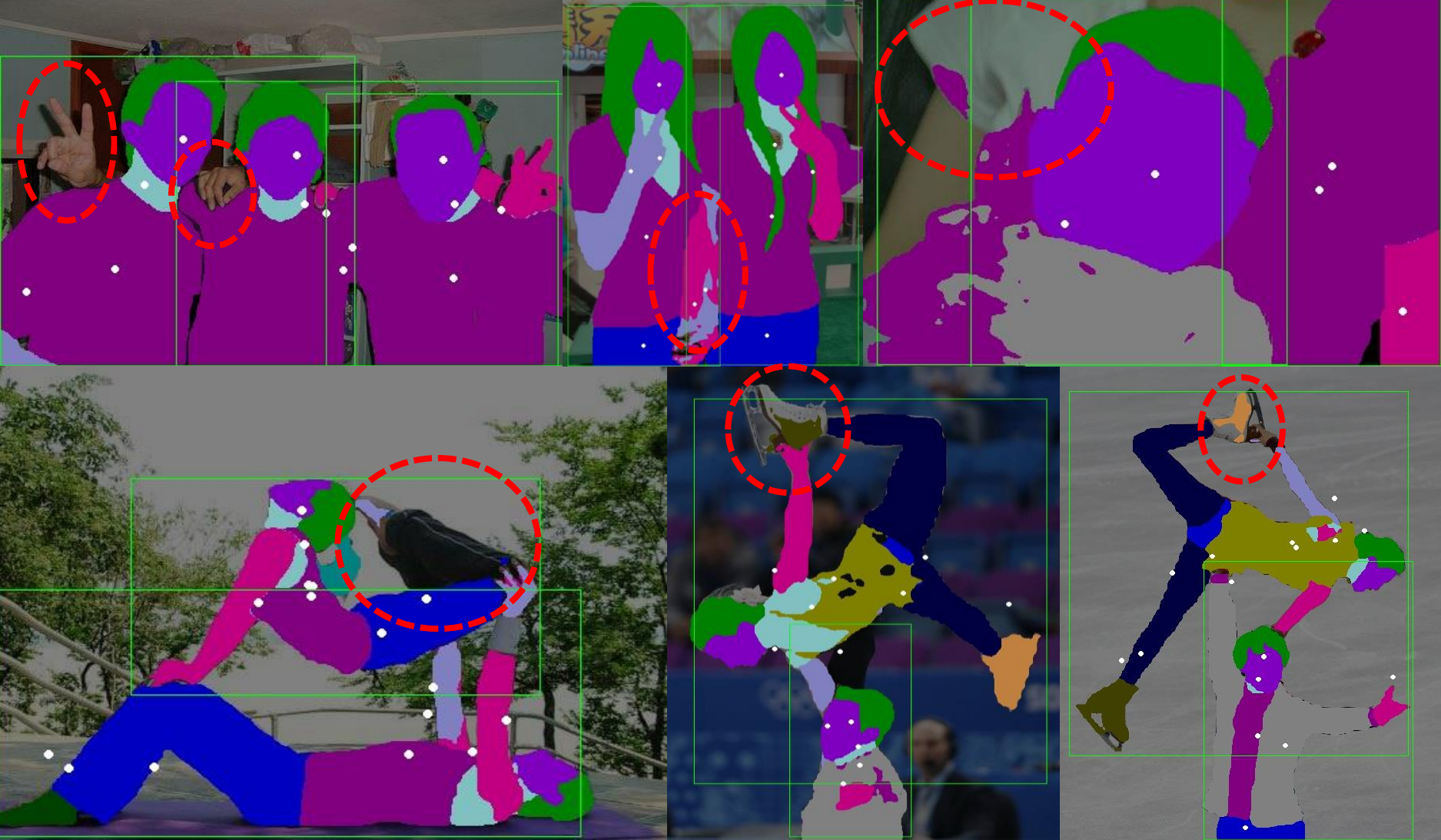}
\caption{Failure cases of the proposed RepParser on CIHP \textit{val}. The red circles spot the region that can not be parsed. The white dots are estimated representative parts by RepParser.}
\label{fig:failure_case}
\end{figure*}

\subsubsection{More Qualitative results.} More qualitative results of RepParser are shown in Fig.~\ref{fig:quality_result}. RepParser can well handle many challenging scenes with occlusions, scale variations, \textit{etc}. Besides, the estimated centers of representative parts reflect salient parts of each person instance. 

\subsubsection{More Failure cases.} Some failure cases of our proposed RepParser are shown in Fig.~\ref{fig:failure_case}. From the visualization results, RepParser can not parse some regions with extreme cases, such as the confusing part regions and dramatic pose variations. We can observe that RepParser fails to distinguish some part regions from other person instances (seen in column 1). Moreover, RepParser can not parse some regions due to the dramatic pose change (seen in column 2). Generalizing from these cases, we can find that each failure case of the part region is significantly interfered by other part region or other person instances. To precisely parse these cases, a method must carefully consider the rich details of the person instance and generate a more discriminative feature representation. We hope that our findings can inspire more research on multiple human parsing.
\end{document}